\documentclass[letterpaper]{article} 
\usepackage{aaai2026}  
\usepackage{times}  
\usepackage{helvet}  
\usepackage{courier} 
\usepackage[hyphens]{url}  
\usepackage{graphicx} 
\usepackage{natbib}  
\usepackage{caption} 
\usepackage{algorithm}
\usepackage{algorithmic}
\usepackage{newfloat}
\usepackage{listings}
\usepackage{amsmath}
\usepackage{booktabs}  
\usepackage{multirow}  
\usepackage{colortbl}  
\usepackage{xcolor}    
\usepackage{amssymb}    
\usepackage{placeins}
\usepackage{graphicx} 
\usepackage{makecell}  
\usepackage{tabularx}

\DeclareCaptionStyle{ruled}{labelfont=normalfont,labelsep=colon,strut=off} 
\floatstyle{ruled}
\newfloat{listing}{tb}{lst}{}
\floatname{listing}{Listing}
\title{Length Matters: Length-Aware Transformer for Temporal Sentence Grounding}

\author{
    Yifan Wang\textsuperscript{\rm 1},
    Ziyi Liu\textsuperscript{\rm 1},
    Xiaolong Sun\textsuperscript{\rm 2},
    Jiawei Wang\textsuperscript{\rm 1},
    Hongmin Liu\textsuperscript{\rm 1}
    }
\affiliations{
    \textsuperscript{\rm 1} School of Intelligence Science and Technology, University of Science and Technology Beijing\\
    \textsuperscript{\rm 2} National Key Laboratory of Human-Machine Hybrid Augmented Intelligence, National Engineering Research Center for Visual Information and Applications, and Institute of Artificial Intelligence and Robotics, Xi'an Jiaotong University\\[0.3em]
    \texttt{yifanwanglingf@google.com, liuziyi@ustb.edu.cn, sunxiaolong@stu.xjtu.edu.cn, m202420939@xs.ustb.edu.cn, hmliu\_82@163.com}
}

\usepackage{bibentry}

\begin{document}

\nocopyright

\maketitle 

\begin{abstract}
Temporal sentence grounding (TSG) is a highly challenging task aiming to localize the temporal segment within an untrimmed video corresponding to a given natural language description. 
Benefiting from the design of learnable queries, the DETR-based models have achieved substantial advancements in the TSG task. 
However, the absence of explicit supervision often causes the learned queries to overlap in roles, leading to redundant predictions. 
Therefore, we propose to improve TSG by making each query fulfill its designated role, leveraging the length priors of the video-description pairs. 
In this paper, we introduce the Length-Aware Transformer (LATR) for TSG, which assigns different queries to handle predictions based on varying temporal lengths. 
Specifically, we divide all queries into three groups, responsible for segments with short, middle, and long temporal durations, respectively. 
During training, an additional length classification task is introduced. Predictions from queries with mismatched lengths are suppressed, guiding each query to specialize in its designated function. 
Extensive experiments demonstrate the effectiveness of our LATR, achieving state-of-the-art performance on three public benchmarks. Furthermore, the ablation studies validate the contribution of each component of our method and the critical role of incorporating length priors into the TSG task.
\end{abstract}

\begin{figure}[!t]
  \centering
  \includegraphics[width=\linewidth]{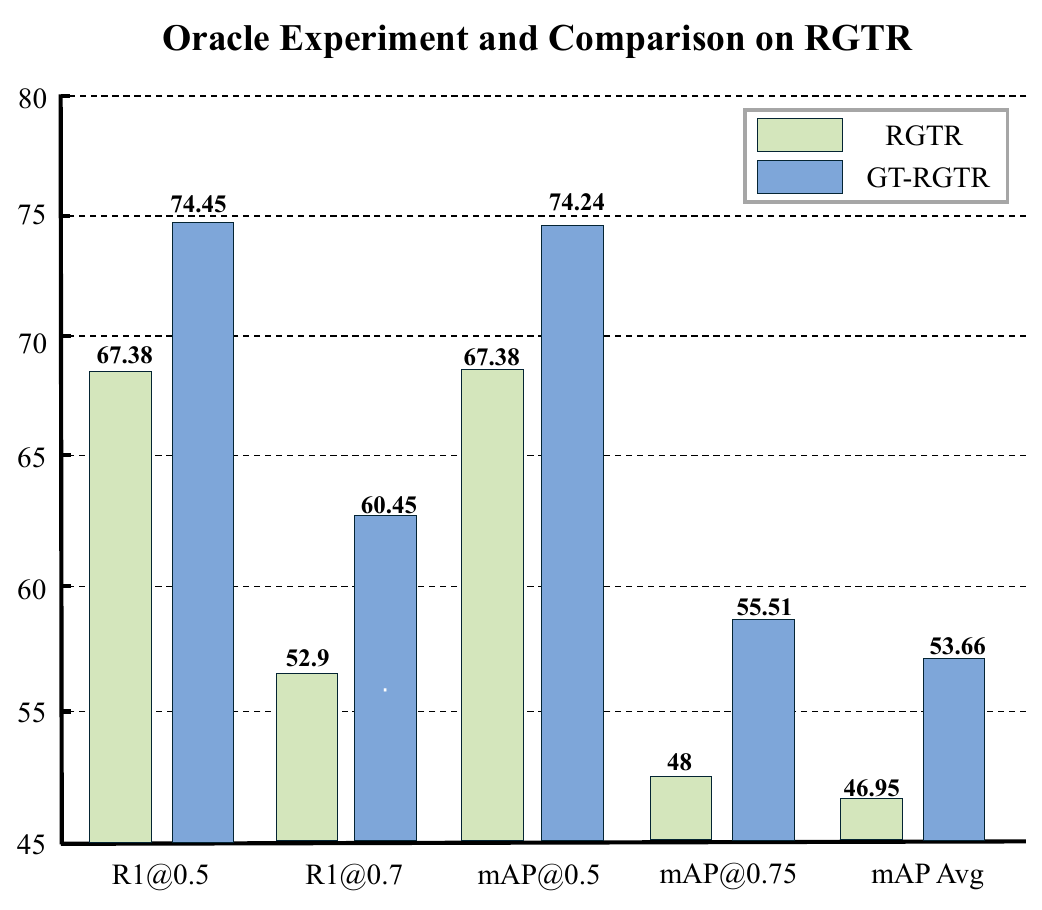}
  \caption{Comparison of the RGTR and RGTR with all length specificity in QVHighlights validation split. Simply incorporating length specificity leads to a significant improvement in model performance. Therefore, length matters.}
  \label{fig1}
\end{figure}

\section{Introduction}
Temporal Sentence Grounding (TSG) task has emerged as a significant research hotspot, which aims to localize the temporal segment within an untrimmed video corresponding to a given natural language description. 
Pioneering research mostly relied on selecting from dense predefined dense proposals\cite{gao2017tall, zhang2020learning} to obtain the target temporal segments.
Recently, the advent and success of the detection transformer (DETR)\cite{carion2020end} have driven the exploration of DETR-based approaches for the TSG task.
Leveraging the advantages of multiple learnable queries, DETR-based approaches have achieved great leaps forward, setting a new benchmark for performance and effectiveness\cite{moon2023query, xiao2024bridging}. 

However, the absence of explicit guidance often causes the learned queries to overlap in roles, leading to redundant predictions. 
We argue that making each query fulfill its designated role, is the key of addressing these limitations.
To achieve this, we leverage the length priors of the video-description pairs to provide additional supervision of the learnable queries during training. 
Intuitively, the duration of the segments is closely related to the content, such as the type of actions.We simply guide the model to learn the mapping from short/middle/long lengths to corresponding queries, enabling each query to fulfill its specific role and reduce overlapping predictions.We assume that the length priors are beneficial to the TSG task, by guiding each query fulfill its designated role.

To prove our assumption, an oracle experiment is carried out as shown in Fig.~\ref{fig1}. 
We directly provide the length labels to the model to validate the theoretical upper bound of leveraging length priors, which significantly outperforms the current SOTA performance.
Motivated by this, we propose the Length-Aware Transformer (LATR) for TSG, which assigns different queries to handle predictions based on varying temporal lengths.
Besides, the following decoder is accordingly adjusted as Length Aware Decoder, employing clustering technique to initial length roles for anchor parts of queries. 
However, such a suppression-only pipeline will make wrong predictions if the input pairs are misclassified. 
To compensate it, we proposed Low-Quality-Residual Masking to handle these misclassifications. 
The length tokens are evaluated before classification, and the quality scores are obtained, indicating whether a sample is suit for the length classification.
Residuals of the low-quality samples are filtered, disabling the suppression.
\begin{figure}[t]
  \centering
  \includegraphics[width=\linewidth ]{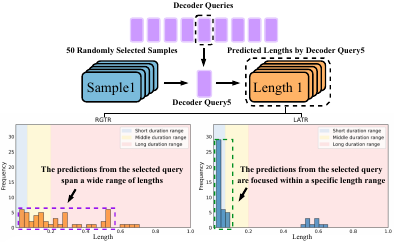}
  \caption{The degree of concentration in lengths predicted by a Single Decoder query. We select a Decoder Query and use it to predict the lengths for 50 randomly selected samples. Then, we generate a histogram showing the frequency distribution of these predicted lengths. The x-axis denotes the normalized moment span length, The y-axis denotes the frequency of the lengths. In comparison, the lengths predicted by RGTR query are relatively dispersed, whereas the lengths predicted by LATR query are more concentrated.}
  \label{fig2}
\end{figure}

Extensive experiments on three TSG benchmarks demonstrate the effectiveness of LATR framework. 
As Shown in Fig.~\ref{fig2}, compared with previous works, the learned queries of our LATR focus on the specific range of length, indicating our learned queries indeed fulfill their length roles. 
In addition, the oracle experiment reveals that the length priors are critical for the TSG task, which has not been fully explored yet.
Our contributions are summarized as follows:
(1) We propose a simple yet effective Length-Aware Transformer (LATR) for TSG, where an additional length classification task assigns short, middle, or long length roles to queries.
(2) We propose a Query-Length Interaction module that converts length classification outputs into residual suppression signals, effectively reducing the impact of mismatched queries.
(3) We design the Low-Quality-Residual Masking to handle the samples with wrong length predictions. 
(4) Facilitated by the proposed module and strategy, our LATR can learn queries focus on specific length roles, achieving SOTA performances on  three TSG benchmarks.

\section{Related Work}
\textbf{Temporal Sentence Grounding.} 
Temporal Sentence Grounding aims to localize the temporal segment within an untrimmed video corresponding to a given natural language description. 
It was first proposed by Gao et al.\cite{gao2017tall}. Compared to general video understanding tasks like VideoQA\cite{9805683,jiang2024ijcv,jiang2021x} or localization \cite{9428587,zhu2022learning,liu2025bridge} tasks, TSG requires precise temporal localization rather than coarse semantic reasoning, making it inherently more challenging.
Early methods relied on proposal-based and proposal-free approaches, but neither was able to significantly improve model performance. 
Proposal-based methods\cite{liu2018cross, wang2022negative, zhang2020learning, sun2022you} generate a large set of dense proposals in advance and rank them based on their similarity with the description. 
Clearly, this approach lacks flexibility and results in considerable redundancy and computational overhead. 
Proposal-free methods\cite{lu2019debug, chen2020rethinking, yang2022entity} directly predict the start and end boundaries of the target moment to avoid relying on a large number of predefined proposals.
However, due to immature feature fusion techniques and boundary decoding strategies, their performance remains suboptimal. 
The recent success of Detection Transformer\cite{carion2020end} has inspired researchers to incorporate Transformer into the TSG framework\cite{lei2021detecting, liu2022reducing, sun2025moment, lee2024bam}. 
DETR-based methods introduce an end-to-end paradigm to eliminate the need for predefined proposals and handcrafted techniques.
They leverage the Transformer encoder as a powerful tool for high-quality multimodal feature fusion, and use learned queries as efficient decoders for target moment localization.
This effectively addresses the limitations of proposal-based and proposal-free methods, significantly boosting model performance and demonstrating strong adaptability and momentum through rapid iteration.\\
\noindent \textbf{Anchor-based DETR.} DETR provides a set of powerful learnable queries to decode the predicted content from features (e.g., spatial extents, temporal center positions of video moments, moment lengths, etc.). However, without explicit guidance, these queries struggle to clearly establish their roles, leading to ambiguous specialization, unstable convergence, and degraded performance—particularly for TSG where role clarity (e.g., center vs. length) is crucial. To address this, a line of work injects anchors into DETR to endow queries with priors and stronger inductive biases, assigning roles and guiding learning more effectively \cite{zhu2020deformable, shi2022motion, wang2022anchor, liu2022dab, zhang2022dino}. In practice, such anchors serve as reference points that shrink the search space and stabilize early decoding. The first to adopt an anchor-based design for TSG was RGTR \cite{sun2024diversifying}, which derives anchors via clustering and adds them to the queries as guidance. This eases optimization but still leaves role assignment largely implicit. Nevertheless, we observe that simply adding anchors yields only weak supervision: query roles may drift across samples and training stages, making it hard to consistently maintain clear responsibilities in practice. To overcome this limitation, we propose a data-driven query constraint that explicitly enforces role identification and preserves it throughout training and inference, enabling queries to remain specialized and stable.

\newpage

\begin{figure*}[t]
  \centering
  \includegraphics[width=\textwidth]{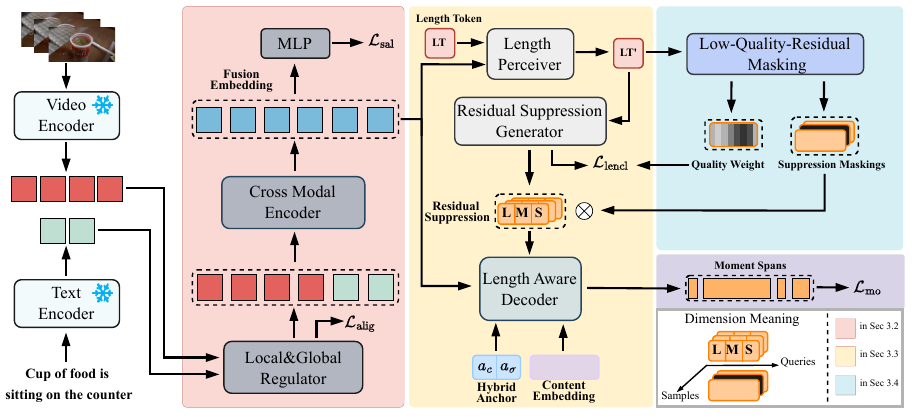}
  \caption{The overall architecture of LATR. It consists of three main components: Feature Extraction and Fusion (in Sec 3.2), Query-Length Interaction Module (in Sec 3.3) and Low-Quality-Residual Masking (in Sec 3.4)}
  \label{fig3}
\end{figure*} 

\section{Method}
\subsection{3.1 Overview}

Given an untrimmed video $\nu =\,\,\left\{ v_{t} \right\} _{t=1}^{L}$ with $L$ frames and a given natural language description $\tau =\,\,\left\{ t_{n} \right\} _{n=1}^{N}$ with N words, TSG aims to localize the moment span $m = (m_{c}, m_{\sigma})$, where $m_{c}$ represent the center time of moment span, and $m_{\sigma}$ represent the duration length of moment span.

The overall architecture of LATR is shown in Fig.~\ref{fig3}.
In general, LATR consists of three parts: Multimodal Feature Extraction and Feature Fusion, Query-Length Interaction Module, and Low-Quality-Residual Masking. 
Firstly, videos and descriptions are processed by the Multimodal Feature Extraction and Fusion Module for feature extraction, alignment, and fusion. 
Secondly, the fusion embedding enters the Query-Length Interaction Module to apply query constraints and predict moment spans and categories. 
Specificity, a Length Token (LT) captures length-specific information, performs classification, and generates Residual Suppression (RS), which is then used by the Length Aware Decoder to guide queries and make final predictions. 
Lastly, the Low-Quality-Residual Masking refines the generation of residual suppression and its interaction to suppress parts of residuals selectively.

\subsection{3.2 Features Extraction and Fusion}

Following previous methods\cite{moon2023query, li2024momentdiff}, the input video and description are fed into pre-trained models to extract video features  $F_{v}\in\mathrm{R}^{L\times d_{v}}$ and textual features $F_{t}\in\mathrm{R}^{N\times d_{t}}$, where $L$ represents the number of clips and $d_{v}$ represents the dimension of visual features; $N$ represents the number of words and $d_{t}$ represents the dimension of textual features.
The video features $F_{v}$ and textual features $F_{t}$ are projected into the common multimodal space using multi-layer perceptrons (MLP) to get the corresponding features $\overline{F}_{v}\in\mathrm{R}^{L\times D}$ and $\overline{F}_{t}\in\mathrm{R}^{N\times D}$, where $D$ is the transformer embedding dimension. 
Then these features are fed into a Local\&Global Regulator\cite{li2021align, sun2024tr}to align.
An additional alignment loss $\mathcal{L}_{\mathrm{alig}}$ is employed to facilitate the alignment between videos and textual descriptions.

After these, the video features and textual features are fed into a Cross Modal Encoder to perform feature fusion and finally obtain the Fusion Embedding $F$.
Following previous methods, the Cross Modal Encoder is imposed by saliency score constraints $\mathcal{L}_{\mathrm{sal}}$\cite{moon2023query}.

\subsection{3.3 Query-Length Interaction Module}
Data-driven design lies at the heart of our approach. 
We argue that the model should actively learn the target moment’s length specificity from each sample and leverage it to guide subsequent processing. 
To achieve this, the Length Perceiver captures length specificity and generates length prediction, which is then transformed by the Residual Suppression Generator into residual suppression. 
Additionally, we revisit the decoder structure and offer a new perspective: moment span prediction involves learning length specificity, albeit in a more explicit manner. 
These two similar processes can complement and reinforce each other.
To this end, we introduce the Length Aware Decoder to realize such interaction.

\begin{figure}[ht]
  \centering
  \includegraphics[width=\linewidth]{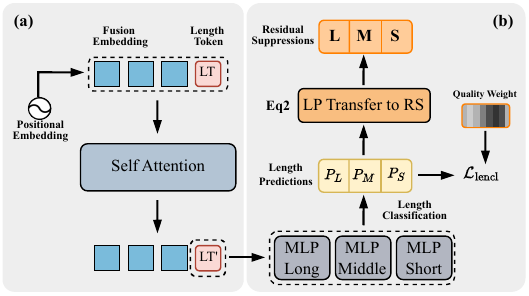}
  \caption{(a) The structure of Length Perceiver. It uses self-attention to pool the target moment’s length specificity from the sample into the Length Token. (b) The structure of Residual Suppression Generator. It performs a classification task through three MLPs, decoding the length specificity encoded in the length token into a length prediction, which is then transformed into residual suppression.}
  \label{fig4}
\end{figure}

\noindent\textbf{Length Prediction.}\ 
The structure of Length Perceiver and Residual Suppression Generator is shown in Fig.~\ref{fig4}. Motivated by the class embedding of Vision Transformer\cite{dosovitskiy2020image}, we utilize self-attention to pool the length-specific information from the Fusion Embeddings $F$ into the length token (LT), formulated as:
\begin{equation}\label{eq1}
  LT^{'},\  F^{'} = Split(Self\mathrm{-}Attention(PE(LT, F))
\end{equation}
where length token $\mathrm{LT^{'}}$ learns the length specificity of the target moment. 
Then, a classification task is performed on the $\mathrm{LT^{'}}$. By feeding the $\mathrm{LT^{'}}$ into three MLP classification heads, we can decode the length specificity as three probability scores, which is the Length Prediction. Specifically, Length  Prediction $= (P_L, P_M, P_S)$, which reflect the likelihood that the sample contains a long/middle/short moment. 

\noindent\textbf{Residual Suppression.} Once the Length Prediction are obtained, the next step is to convert them into constraint guidance information, thereby enabling the match between the sample's length specificity and the queries in the decoder to provide guidance based on length specificity. 
Therefore, the Length Prediction is fed into the Residual Suppression Generator to be transformed into Residual Suppression (RS), which directly interacts with the queries to impose constraint-based guidance.
Specifically, the generation process of RS is formulated as follows:

\begin{equation}\label{eq2}
\begin{aligned}
  L  &= \tau - P_L \\
  M  &= \tau - P_M \\
  S  &= \tau - P_S \\
  RS &= (L, M, S)
\end{aligned}
\end{equation}

\noindent where $\tau$ is a hyperparameter representing the classification confidence threshold, if the probability exceeds this confidence threshold, the sample is considered to contain a target moment of the corresponding length; L, M, and S are all scalar values.

Once the residual suppression (RS) is generated, it can be applied to the queries to impose constraints. 
For queries whose length roles are inconsistent with the predicted length specificity, RS suppresses their content embeddings; conversely, for those with aligned length roles, the suppression is minimal or bypassed. 
This selective suppression mechanism indirectly enhances the utility of consistent queries by reducing interference from mismatched ones.

Specifically, following human intuition, we divide the queries into Long, Middle, and Short categories and extract the content embedding of these queries, denoted as $E_{Long}, E_{Middle}, E_{Short} \in \mathrm{R}^D$.
The residual suppression is applied to the content embedding as the formulation:

\begin{equation}\label{eq3}
\setlength{\arraycolsep}{1.0ex} 
\begin{array}{l @{\;} c @{\;} l @{\;} c @{\;} l @{\;} c @{\;} l}
E_{Long}   & = & E_{Long}   & + & L & \times & E_{Long} \\
E_{Middle} & = & E_{Middle} & + & M & \times & E_{Middle} \\
E_{Short}  & = & E_{Short}  & + & S & \times & E_{Short}
\end{array}
\end{equation}

Through residual suppression, queries aligned with the sample's length specificity retain their content embedding, while mismatched ones are suppressed.

\begin{figure}[!t]
  \centering
  \includegraphics[width=\linewidth]{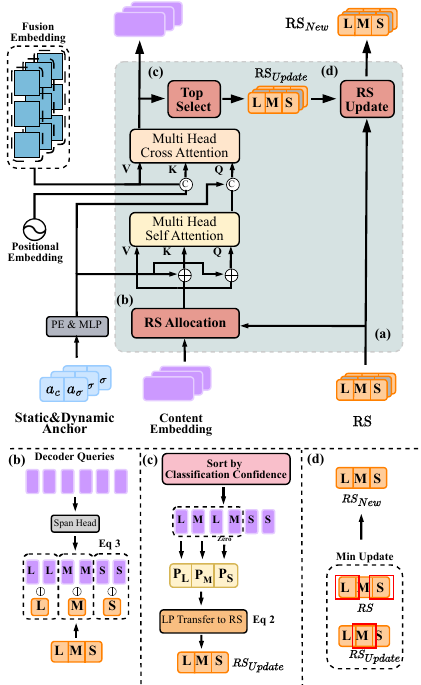}
  \caption{(a) The structure of Length Aware Decoder. We argue that the process of predicting moment spans essentially reflects the acquisition of the target moment’s length specificity. (b) The structure of RS Allocation, the RS is allocated to the corresponding query. (c) The structure of Top Select, the RS used for update is generated here. (d) The structure of RS Update, the new RS is generated by selecting the smaller values between the previous RS and the updated RS.
}
  \label{fig5}
\end{figure}

\noindent \textbf{Length Aware Decoder.}
\ Once the Fusion Embedding \( F \) and residual suppression \( \mathrm{RS} \) are ready, they are fed into the decoder to predict the moment span.
We revisit the decoder architecture and the moment span prediction process from a new perspective of length specificity, aiming to observe how the sample’s length characteristics influence the decoding. 
This analysis reveals several interesting findings.

In fact, in the TSG task, the moment spans predicted by the decoder also include length specificity, reflecting precise numerical lengths. 
Moreover, the foreground confidence scores for predicted lengths closely resemble the probabilistic outputs discussed earlier for short, medium, and long moments. 
This suggests that decoder layers can serve as an alternative Length Perception module, generating a variant of residual suppression.
This variant dynamically refines the residual suppression from the length token (LT).

The structure of Length Aware Decoder is shown as Fig.~\ref{fig5}.
Following previous methods\cite{sun2024diversifying}, we employ k-means clustering to generate anchors, which are then projected through an MLP and positionally encoded to form the positional embeddings of queries. Essentially, this initialization provides the queries with an initial length role. The content embedding of queries is zero-initialized.

\subsection{3.4 Low-Quality-Residual Masking}
As discussed in Sec 3.3, LATR is inherently data-driven, making accurate length classification and residual suppression essential. 
Since the method relies on residual suppression, it becomes overly aggressive and fragile—incorrect suppression can significantly degrade performance. 
To address this, we propose the Low-Quality-Residual Masking, which focuses on better utilizing, rather than refining, residual suppression. 

\begin{figure}[t]
  \centering
  \includegraphics[width=\linewidth]{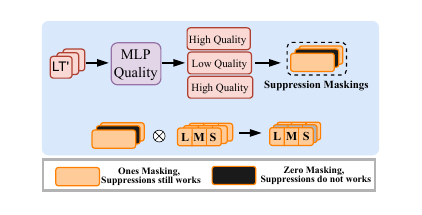}
  \caption{ The process of Low-Quality-Residual Masking. A Ones Masking is generated for high-quality Length Tokens, so the corresponding RS is retained and still work. Accordingly, a Zero Masking is generated, so the corresponding RS is masked out and no longer work.}
  \label{fig6}
\end{figure}

To address the misclassifications and incorrect residual suppression caused by these complex samples, we feed the length token into an MLP-based quality evaluator, MLP-Quality, to obtain quality scores $Qs \in \mathrm{R}^{bs}$, which are used to determine whether a sample is complex.
Based on $Qs$, we then generate Suppression Maskings $\in \mathrm{R}^{bs \times N_q \times 1}$ to mask out the residual suppression generated by complex samples, such that these samples are no longer regulated by residual suppression. 
Fig.~\ref{fig6} shows the detailed structure of this strategy.
The proposed strategy also proves effective in handling length annotation noise in the dataset, where Low-Quality-Suppression Masking selectively suppresses anomalous samples to prevent them from interfering with training.

Furthermore, misclassified samples not only lead to incorrect residual suppression but also interfere with the classification loss, affecting its optimization. 
To mitigate this issue, we adopt a weighted cross-entropy loss calculation, where $Qs$ is used as the weighting factor.

To prevent the model from collapsing into a trivial solution—e.g., assigning a significantly high $Qs$ to one sample and very low $Qs$ to others, we adopt a combination of the standard mean loss $\mathcal{L}{mean}$, the weighted loss $\mathcal{L}{weight}$, and median loss $\mathcal{L}{median}$ to train the classification module. 

The median-difference loss is defined as the absolute difference between the median and minimum of $Qs$, which reflects the severity of the trivial prediction problem.
The total classification loss is defined as: 

\begin{equation}\label{eq4}
    \mathcal{L}_{lencl} =  \mathcal{L}_{mean} +  \mathcal{L}_{weight} + \mathcal{L}_{median}
\end{equation}

Additionally, we propose a \textbf{TopK-Save} strategy, which interacts with the Length Aware Decoder. 
By simply retaining a small number of high-confidence Decoder Queries that would otherwise be suppressed, this enables the model to perform self-correction through the Length Aware Decoder.

\subsection{3.5 Training Objectives}
LATR consists of four objective loss components: moment loss $\mathcal{L}_{mo}$, saliency loss $\mathcal{L}_{sal}$, alignment loss $\mathcal{L}_{alig}$ and additional length classification loss $\mathcal{L}_{lencl}$. The overall objective is defined as:
\begin{equation}\label{eq5}
    \mathcal{L}_{} = \mathcal{L}_{mo} + \lambda_{sal} \times \mathcal{L}_{sal} + \lambda_{alig} \times \mathcal{L}_{alig} + \lambda_{lencl} \times \mathcal{L}_{lencl}
\end{equation}

\noindent where $\lambda_{*}$ are the balancing parameters. It is important to note the values of $\lambda_{sal}$ and $\lambda_{alig}$ are determined based on extensive experimentation in the TSG community and are considered optimal, thus requiring no further discussion. In addition, we investigate the impact of different values of $\lambda_{lencl}$ on model performance through ablation studies in Sec 4.4.

\section{Experiment}

\begin{table*}[h]
\centering
\caption{Performance Comparison on QVHighlights \textit{val} and \textit{test} splits.}
\renewcommand{\arraystretch}{1.0}
\begin{tabularx}{\linewidth}{l *{10}{>{\centering\arraybackslash}X}}
\toprule
& \multicolumn{5}{c}{val} & \multicolumn{5}{c}{test}\\
\cmidrule(lr){2-6} \cmidrule(lr){7-11} 
Method & \multicolumn{2}{c}{R1} & \multicolumn{3}{c}{mAP} & \multicolumn{2}{c}{R1} & \multicolumn{3}{c}{mAP}\\
\cmidrule(lr){2-3} \cmidrule(lr){4-6} \cmidrule(lr){7-8} \cmidrule(lr){9-11} 
& @0.5 & @0.7 & @0.5 & @0.75 & Avg. & @0.5 & @0.7 & @0.5 & @0.75 & Avg.\\ 
\midrule 
    M-DETR~\cite{lei2021detecting}          & 53.94 & 34.84 & --    & --    & 32.20 & 52.89 & 33.02 & 54.82 & 29.40 & 30.73 \\
    UMT~\cite{liu2022umt}                   & 60.26 & 44.26 & --    & --    & 38.59 & 56.23 & 41.18 & 53.83 & 37.01 & 36.12 \\
    QD-DETR~\cite{moon2023query}            & 62.68 & 46.66 & 62.23 & 41.82 & 41.22 & 62.40 & 44.98 & 62.52 & 39.88 & 39.86 \\
    UniVTG~\cite{lin2023univtg}             & 59.74 &  --   &  --   &  --   & 36.13 & 58.86 & 40.86 & 57.60 & 35.59 & 35.47 \\
    TR-DETR~\cite{sun2024tr}                & 67.10 & 51.48 & 66.27 & 46.42 & 45.09 & 64.66 & 48.96 & 63.98 & 43.73 & 42.62 \\
    TaskWeave~\cite{yang2024task}           & 64.26 & 50.06 & 65.39 & 46.47 & 45.38 &  --   &  --   &  --   &  --   &  --   \\
    UVCOM~\cite{xiao2024bridging}           & 65.10 & 51.81 &  --   &  --   & 45.79 & 63.55 & 47.47 & 63.37 & 42.67 & 43.18 \\
    CG-DETR~\cite{moon2023correlation}      & 67.35 & 52.06 & 65.57 & 45.73 & 44.93 & \underline{65.43} & 48.38 & 64.51 & 42.77 & 42.86 \\
    RGTR~\cite{sun2024diversifying}         & \underline{67.68} & \underline{52.90} & \underline{67.38} & \underline{48.00} & \underline{46.95} &
                                              \textbf{65.50} & \underline{49.22} & \underline{67.12} & \underline{45.77} & \underline{45.53} \\
    \rowcolor{gray!15}
    \textbf{LATR (Ours)}                    & \textbf{67.74} & \textbf{53.16} & \textbf{68.03} & \textbf{49.10} & \textbf{48.16} &
                                              65.30 & \textbf{49.35} & \textbf{67.33} & \textbf{46.12} & \textbf{46.07} \\
\bottomrule 
\end{tabularx}

\label{t1}
\end{table*}

\begin{table*}[!ht]
\tabcolsep=0.42cm
\centering
\caption{Performance comparison on TACoS and Charades-STA.}
\begin{tabularx}{\linewidth}{l *{6}{>{\centering\arraybackslash}X}}
    \toprule
    \multirow{2}{*}{Method} & 
    \multicolumn{3}{c}{\textbf{Charades-STA}} & 
    \multicolumn{3}{c}{\textbf{TACoS}} \\
    \cmidrule(lr){2-4} \cmidrule(lr){5-7}
    & R1@0.5 & R1@0.7 & mIoU  & R1@0.5 & R1@0.7 & mIoU \\
    \midrule
    M-DETR~\cite{lei2021detecting}    & 52.07 & 30.59 & 45.54 & 24.67 & 11.97 & 25.49\\
    UMT (Liu et al. 2022)             & 48.31 & 29.25 & -- & -- & -- & -- \\
    QD-DETR~\cite{moon2023query}      & 57.31 & 32.55 & -- & -- & -- & --\\
    UniVTG~\cite{lin2023univtg}       & 58.01 & 35.65 & 50.10 & 34.97 & 17.35 & 33.60\\
    TR-DETR~\cite{sun2024tr}          & 57.61 & 33.52 & -- & -- & -- & --\\
    CG-DETR~\cite{moon2023correlation}   & \underline{58.44} & \underline{36.34} & 50.13 & 39.61 & 22.23  & 70.43 \\
    RGTR~\cite{sun2024diversifying}   & 57.93 & 35.16 & \underline{50.32} & \underline{40.31} & \underline{24.32} & \underline{37.44}\\
    \rowcolor{gray!15}
    \textbf{LATR (Ours)}    & \textbf{59.54} & \textbf{37.33} & \textbf{50.88} & \textbf{40.62} & \textbf{25.37} & \textbf{37.68}\\
    \bottomrule
\end{tabularx}

\label{t2}
\end{table*}

\subsection{4.1 Datasets and Metrics}
\textbf{Datasets.} \ \  We evaluate LATR on three temporal sentence grounding benchmarks, including the QVHighlights\cite{lei2021detecting}, Charades-STA\cite{gao2017tall} and TACoS\cite{regneri2013grounding}. 
QVHighlights spans various themes, and multiple target moments of varying lengths typically exist within a single video-description query pair. 
Charades-STA comprises intricate daily human activities, and TACoS mainly showcases long-form videos focusing on culinary activities. For these two datasets, only a single target moment exists within each video-query pair. \\
\textbf{Metrics.}  \ \  We adopt the Recall@1 (R1) under the IoU thresholds of 0.3, 0.5, and 0.7.
Since QVHighlights contains multiple ground-truth moments per sentence, we also report the mean average precision (mAP) with IoU thresholds of 0.5, 0.75, and the average mAP over a set of IoU thresholds [0.5: 0.05: 0.95].
For Charades-STA and TACoS, we compute the mean IoU of top-1 predictions.

\subsection{4.2 Implementation Details}
For the categorization of target moment lengths, in QVHighlights, by following the empirical insights provided by previous studies \cite{diwan2023zero}, we use the \textbf{absolute length} of the moment for classification: moments with a duration of 0–10 seconds are categorized as short moments, 10–30 seconds as middle moments, and 30–150 seconds as long moments. 
For Charades-STA, we use the \textbf{normalized length} of the moment for classification, moments with 0-0.2 are labeled as short, 0.2-0.302 as middle, and 0.302-1 as long.
The same classification rules are applied to TACoS, where 0-0.045 corresponds to short moments, 0.045-0.1 to middle moments, and 0.1-1 to long moments. \\
\indent As shown in Tab.~\ref{t8}, our ablation study demonstrates that under the principle of equal division, the specific way of length partitioning does not significantly affect model performance. Therefore, what truly matters is performing the length partition itself, rather than the exact method of partitioning. Based on this observation, we propose a general guideline for length division on new datasets: we suggest dividing moment lengths into three categories—short, middle, and long—based on human intuition. To account for varying video durations, we recommend using normalized lengths. The division can be achieved by approximately splitting the training samples into three groups with balanced quantities.\\
\indent Following previous methods\cite{moon2023query}, we use SlowFast\cite{feichtenhofer2019slowfast} and CLIP\cite{radford2021learning} to extract visual features and CLIP to extract textual features.
We set the embedding dimension D to 256. The number of queries and anchor pairs is set to 20 for QVHighlights, 10 for Charades-STA and TACoS.
The balancing parameters are set as: $\lambda_{alig} = 0.3$, $\lambda_{sal}$ is set as 1 for QVHighlights, 4 for Charades-STA and TACoS,and $\lambda_{lencls}$ is set as 1 for QVHighlights, 3 for Charades-STA and TACoS.The Top-Select is set as 4 for QVHighlights and Charades-STA, 3 for TACoS. 
The TopK-Save is set as 3 for QVHighlights, 1 for Charades-STA and TACoS. 
The $\tau$ is set as 0.5. We train all models with batch size 32 using the AdamW optimizer with weight decay 1e-4. The learning rate is set to 1e-4.

\subsection{4.3 Performance Comparison}
As shown in Tab.~\ref{t1}, we compare LATR to previous methods on QVHighlights validation and test splits. 
Our method achieves new state-of-the-art performance on almost all metrics of validation split and test split. The well performance

\newcommand{\tablesep}{\vspace{-2pt}}

\clearpage
\twocolumn[{%
\noindent
\begin{minipage}[t]{\dimexpr0.5\textwidth-0.5\columnsep\relax}\centering

\captionof{table}{Comparison of GFLOPs, Parameters, and mAP$_{\text{avg.}}$ for RGTR and LATR.}
\begin{tabularx}{\linewidth}{>{\centering\arraybackslash}X
                          >{\centering\arraybackslash}X
                          >{\centering\arraybackslash}X
                          >{\centering\arraybackslash}X}
\toprule
Method & GFLOPs & Parameters & mAP$_{\text{avg.}}$ \\
\midrule
RGTR & 153.07 & 9.22M  & 36.95 \\
LATR & 174.47 & 11.65M & 38.16 \\
\bottomrule
\end{tabularx}
\label{t3}

\tablesep

\captionof{table}{Ablation study on the impacts of different Length Token (LT)
Initialization Strategies, Classification Strategies, and Suppression Strategies.}
{\setlength{\tabcolsep}{3pt}%
\begin{tabular*}{\linewidth}{@{\extracolsep{\fill}} c c c c c @{}}
  \toprule
  Method & Changes & R1@0.5 & mAP@0.5 & mAP$_{\text{avg.}}$ \\
  \midrule
  \multirow{2}{*}{LT Initialization}
    & random       & 67.10 & 67.32 & 47.43 \\
    & \textbf{zero}& \textbf{67.74} & \textbf{68.03} & \textbf{48.16} \\
  \midrule
  \multirow{2}{*}{Classification}
    & 1$\times$ 3-way        & 67.68 & 67.43 & 47.31 \\
    & \textbf{3$\times$binary}& \textbf{67.74} & \textbf{68.03} & \textbf{48.16} \\
  \midrule
  \multirow{2}{*}{Suppression}
    & fixed factor  & 67.48 & 67.54 & 47.49 \\
    & \textbf{residual} & \textbf{67.74} & \textbf{68.03} & \textbf{48.16} \\
  \bottomrule
\end{tabular*}}
\label{t5}

\tablesep

\captionof{table}{Comparison of RGTR and LATR on the Length Evaluation Dataset, composed of QVHighlights validation samples that are accurately and perfectly classified by LATR.}
\renewcommand{\arraystretch}{1.0}
\begin{tabularx}{\linewidth}{
  >{\raggedright\arraybackslash}X *{5}{>{\centering\arraybackslash}X}}
\toprule
\multirow{2}{*}{Method} & \multicolumn{2}{c}{R1} & \multicolumn{3}{c}{mAP} \\
\cmidrule(lr){2-3}\cmidrule(lr){4-6}
& @0.5 & @0.7 & @0.5 & @0.75 & Avg. \\
\midrule
RGTR  & 70.33 & 54.07 & 55.84 & 33.67 & 33.54 \\
LATR  & 74.64 & 55.98 & 60.64 & 33.89 & 35.89 \\
\bottomrule
\end{tabularx}
\label{t7}

\end{minipage}%
\hspace{\columnsep}%
\begin{minipage}[t]{\dimexpr0.5\textwidth-0.5\columnsep\relax}\centering

\captionof{table}{Ablation study on the components of LATR, including the Length
Prediction and Residual Suppression (LP\&RS), Length Aware Decoder (LAD),
Low-Quality-Residual Masking (LQM), and TopK-Save (T-S).}
\centering
\begin{tabular}{@{}c@{\hspace{2pt}}c@{\hspace{2pt}}c@{\hspace{2pt}}c@{\hspace{2pt}}c
                @{\hspace{2pt}}c@{\hspace{2pt}}c@{\hspace{2pt}}c@{}}
  \toprule
  Setting & LP\&RS & LAD & LQM & T\textendash S & R1@0.5 & mAP@0.5 & mAP$_\text{avg.}$ \\
  \midrule
  (a) &  &  &  &  & 67.68 & 67.38 & 46.95 \\
  (b) & \checkmark &  &  &  & 67.29 & 67.41 & 47.52 \\
  (c) & \checkmark & \checkmark &  &  & 67.26 & 67.51 & 47.72 \\
  (d) & \checkmark & \checkmark & \checkmark &  & 67.35 & 67.79 & 47.34 \\
  (e) & \checkmark & \checkmark &  & \checkmark & 67.61 & 67.60 & 47.52 \\
  (f) & \checkmark & \checkmark & \checkmark & \checkmark & \textbf{67.74} & \textbf{68.03} & \textbf{48.16} \\
  \bottomrule
\end{tabular}
\label{t4}

\tablesep

\captionof{table}{Ablation study on the components of length classification loss.}
{\tabcolsep=0.06cm
\begin{tabular*}{\linewidth}{@{\extracolsep{\fill}} c c c c c c c @{}}
  \toprule
  Setting & mean & weight & median & R1@0.5 & mAP@0.5 & mAP$_{avg.}$ \\
  \midrule
  (a) & \checkmark &  &  & 67.23 & 67.41 & 47.48 \\
  (b) &  & \checkmark &  & 67.35 & 67.58 & 47.56 \\
  (c) & \checkmark & \checkmark &  & 67.64 & 67.76 & 47.66 \\
  (d) &  & \checkmark & \checkmark & 67.29 & 67.56 & 47.44 \\
  (e) & \checkmark & \checkmark & \checkmark & \textbf{67.74} & \textbf{68.03} & \textbf{48.16} \\
  \bottomrule
\end{tabular*}}
\label{t6}

\tablesep

\captionof{table}{Ablation study on the length split method.}
\renewcommand{\arraystretch}{1}
\begin{tabularx}{\linewidth}{l *{5}{>{\centering\arraybackslash}X}}
  \toprule
  \multirow{2}{*}{Method} & \multicolumn{2}{c}{R1} & \multicolumn{3}{c}{mAP} \\
  \cmidrule(lr){2-3}\cmidrule(lr){4-6} 
  & @0.5 & @0.7 & @0.5 & @0.75 & Avg. \\
  \midrule
  2 Split & 67.42 & 52.78 & 68.00 & 48.73 & 47.98 \\
  3 Split & 67.74 & 53.16 & 68.03 & 49.10 & 48.16 \\
  4 Split & 67.08 & 52.32 & 68.01 & 48.47 & 47.93 \\
  \bottomrule
\end{tabularx}
\label{t8}

\end{minipage}%
}] %

\noindent demonstrate the effectiveness of queries with clear and consistent length roles.
Tab.~\ref{t2} present comparisons on TACoS and Charades-STA test splits. 
Our method achieves the best performance on TACoS and Charades-STA. \\
\indent Tab.~\ref{t3} presents the additional number of parameters and computational overhead introduced by LATR compared to the baseline. LATR only introduces an extra 2M parameters and incurs approximately 13\% more GFLOPs, resulting in minimal parameter and computation overhead while achieving notable performance improvements. This demonstrates that our proposed method is both simple and effective.

\subsection{4.4 Ablation Study}

\noindent\textbf{Main Ablation.} 
We investigate the effectiveness of each component in LATR. 
In Tab.~\ref{t4}, we report the impact according to  the Length Prediction and Residual Suppression(LP\&RS), Length Aware Decoder(LAD), Low-Quality-Suppression Masking(LQM) and TopK-Save(T-S). 
The results show that each component contributes to performance improvements, and the full model in setting (f) achieves the best performance, improving mAP$_{avg}$ by 1.21.

\begin{figure*}[!t]
  \centering
  \includegraphics[width=\linewidth]{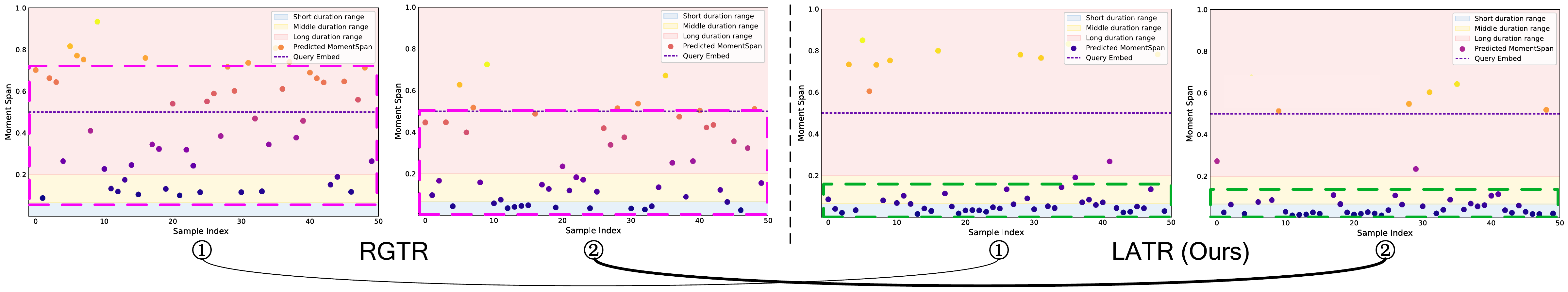}
  \caption{Two Groups of Visualization comparison of moment span predictions for a Single Decoder query over 50 randomly selected samples \textbf{(Following the method mentioned in Fig. 2)} in RGTR and LATR (ours). The x-axis denotes the sample indices, and the y-axis represents the normalized moment span lengths.}
  \label{fig7}

\vspace{2pt} 

\begin{minipage}[t]{0.49\textwidth}
  \vspace{0pt} 
  \centering
  \includegraphics[width=0.8\linewidth]{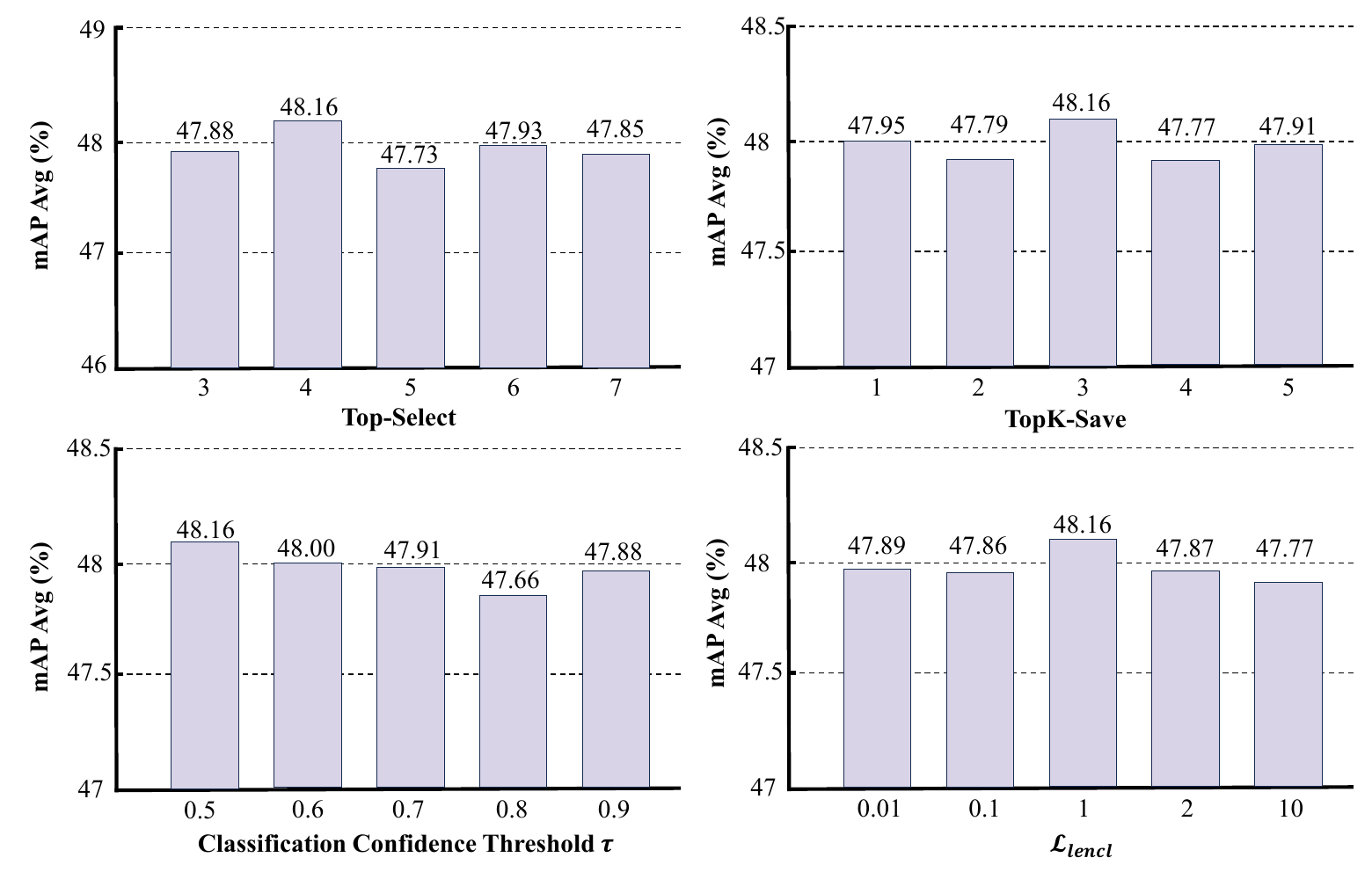}
  \captionsetup{type=figure}
  \captionof{figure}{Ablation study on hyperparameters}
  \label{fig8}
\end{minipage}
\hfill
\begin{minipage}[t]{0.49\textwidth}
  \vspace{0pt} %
  \centering
  \includegraphics[width=\linewidth]{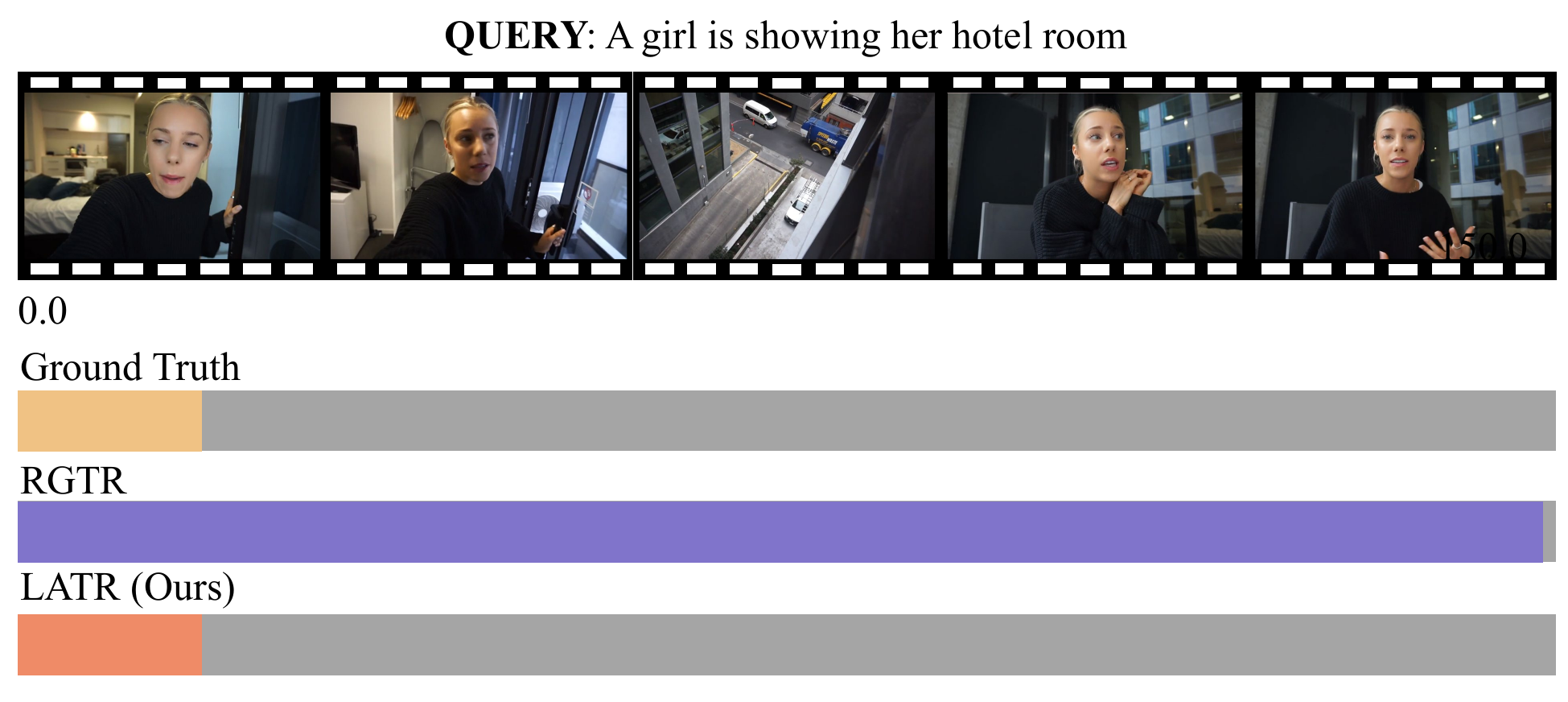}
  \captionsetup{type=figure}
  \captionof{figure}{Illustration of qualitative example}
  \label{fig9}
\end{minipage}

\end{figure*}

\noindent\textbf{Implementation Variants of Key Components.} 
We evaluate different implementation variants of three key components: LT Initialization Strategy, Classification Strategy, and Suppression Strategy. 
Specifically, we experiment with initializing the Length Token using random initialization instead of zero initialization; replacing the three binary classification MLPs with a single three-way classification MLP; and using a pair of fixed scaling factors $S_{positive}\ ,\ S_{negative}$ as an alternative to residual suppression, where the former aims to enhance prominence and the latter is designed for suppression. 
As shown in Tab.~\ref{t5}, All these alternative designs lead to performance degradation, demonstrating that our chosen implementations are the most effective.
More importantly, using fixed scaling factors results in a notable performance drop, further indicating that enhancing the prominence of a query is inherently challenging, with no effective direct solution available. 

\noindent\textbf{Length Classification Loss Function.} We evaluate the contribution of each loss component in the length classification loss function(Eq.~\ref{eq4}) . 
As shown in Tab.~\ref{t6}, using only the default mean cross-entropy loss leads to a drop in model performance, while incorporating additional loss terms based on the quality scores of the Length Token improves the results, demonstrating that each component contributes effectively to the overall optimization. 

\noindent\textbf{Additional Oracle Experiment of Length Significance } To further examine the effect of length information in Temporal Sentence Grounding, we construct an additional Length Evaluation Dataset by extracting all samples from the QVHighlights validation split that are accurately and perfectly classified by LATR. As shown in Tab.~\ref{t7}, LATR significantly outperforms RGTR on this curated set, clearly indicating that the incorporation of length information enhances model performance in TSG. This further substantiates our central claim: length matters.

\noindent\textbf{Length Split Method and Length Split Thresholds.} Under the principle of equal-proportion split, we compare the effects of 2, 3, and 4 length splits on model performance. 
As shown in Tab.~\ref{t8}, the performance differences across these splitting methods are relatively small. 
This indicates that while length splitting is important, the specific way of division is not particularly critical.
We further conduct an ablation study showing that different length division thresholds have little effect on model performance.

\noindent\textbf{Hyperparameter Analysis} We evaluate the impact of different values of hyperparameters. As shown in Fig.~\ref{fig8}, the model performance shows minimal variation across different values of the hyperparameters, demonstrating that the effectiveness of our proposed method does not rely heavily on the specific choice of hyperparameters.

\subsection{4.5 Visualization  Result}
Fig. ~\ref{fig7} visualizes the predicted moment spans of two queries, \textbf{following the method mentioned in Fig.~\ref{fig2}}, set under both RGTR and LATR.
It is clearly observed that the moment spans predicted by LATR are more concentrated within a specific length range, whereas those predicted by RGTR are more widely distributed, indicating less focus.

In Fig.~\ref{fig9}, we present an example on the QVHighlights validation split. When the target moment is short, baseline RGTR predicts it as long, whereas LATR identifies and predicts it as short. This highlights LATR's strong length-awareness for estimating target-moment duration.

\section{Conclusion}

In this paper we proposed LATR, a Length-Aware Transformer for TSG that tackles a core weakness of DETR-style models: queries neither recognize nor consistently maintain their length roles. LATR introduces a data-driven query constraint that captures per-sample length specificity and converts it into residual suppression to guide queries throughout decoding. To improve reliability when length prediction is uncertain, we designed Low-Quality-Residual Masking and TopK-Save, enabling the model to self-correct rather than over-suppress. Across three public benchmarks, LATR achieves state-of-the-art results with minimal overhead (small parameter and GFLOP increases), and ablations verify the contribution of each component as well as the importance of injecting length priors into TSG. Qualitative analyses show that queries under LATR specialize and remain stable, producing concentrated length predictions.

Looking ahead, LATR is a general plug-in that can be integrated into other DETR-based TSG models and related video–language tasks. Our oracle experiment underscores that making length priors explicit is a high-leverage direction, demonstrating both the value and potential of this research. To our knowledge, LATR is the first to structurally encode and operationalize length priors for TSG. The approach remains simple and lightweight—adding only 2M parameters and about 13\% GFLOPs—yet delivers consistent SOTA gains, highlighting a practical path to further progress on temporal sentence grounding. In practice, the same residual-suppression mechanism can be dropped into existing decoders with minimal changes.

\bibliography{aaai2026}

\end{document}